%% file: main.tex
\def\arxiv{}
\documentclass[11pt]{article}

\ifdefined\arxiv
  \usepackage[preprint]{acl}
\else
  \usepackage[review]{acl}
\fi

\usepackage{times}
\usepackage{latexsym}
\usepackage[T1]{fontenc}
\usepackage[utf8]{inputenc}
\usepackage{microtype}
\usepackage{inconsolata}
\usepackage{graphicx}
\usepackage{booktabs}
\usepackage{array}
\usepackage{amsmath}
\usepackage{enumitem}

\newcommand{\pp}{\,\text{pp}}
\newcommand{\ci}[2]{{\footnotesize[$#1$,\,$#2$]}}
\newcommand{\holm}{$^{\dagger}$}

\ifdefined\arxiv
  \newcommand{\codenote}{\footnote{Code: \url{https://github.com/parsa-mz/crtitxer}}}
\else
  \newcommand{\codenote}{\footnote{Code and all artefacts accompany this submission as
    supplementary material.}}
\fi

\title{Prior Audit--Repair Context Shifts LLM Verifier Thresholds Toward Leniency}

\ifdefined\arxiv
  \author{Parsa Mazaheri \\
    University of California, Santa Cruz \\
    \texttt{pmazaher@ucsc.edu} \\\And
    Kasra Mazaheri \\
    Massachusetts Institute of Technology \\
    \texttt{mazaheri@mit.edu}}
\else
  \author{Anonymous ACL submission}
\fi

\begin{document}
\maketitle

\begin{abstract}
Automated checking pipelines increasingly place one language model as the checker and another (or
the same one) as the fixer. We ask whether that wiring changes what the checker reports. Measuring
false alarms on human-verified-correct ProcessBench traces with the present task held byte-identical,
we find that a completed audit\,$\to$\,repair episode already in the model's context lowers false
alarms in \textbf{15 of 15} model\,$\times$\,wording combinations, by $2.8$ to $11.5\pp$ against a
length-matched non-audit control, a $9$ to $25\%$ reduction relative to that control. The direction contradicts
what the accumulated-message literature predicts: an episode whose audit reported an \emph{error}
lowers false alarms \emph{further} still, at \textbf{all five} wordings on the model where that
manipulation lands cleanly, though a negativity asymmetry predicts more flagging. Decomposing the
episode finds repair content and audit verdict complementary: different components carry the effect on
different model families. Signal-detection analysis locates the change in the threshold rather than in
discrimination --- the criterion moves in $15$ of $15$ combinations and survives correction in $13$
while $d'$ survives in none, though the $d'$ test is half as sensitive by construction --- and a hand audit of 50 false alarms finds 82\% simply
wrong, so at this operating point the shift need not be harmful. With reasoning
enabled the effect keeps its relative size on both models tested, and the threshold reading holds
there too.
\end{abstract}

\section{Introduction}
\label{sec:intro}

When a language model checks work --- reviewing code, grading a proof, verifying a reasoning trace
--- it is increasingly wired to something downstream: sometimes a second model repairs what it flags,
sometimes it is handed its own findings and asked to fix them. Choosing between those is treated as an
engineering detail, a question of how to arrange the pipeline rather than of what the checker will
say.\codenote

This paper shows it is not a detail, and that the direction of the effect is not the one the nearest
literature predicts. The practical finding is a caution: \textbf{be wary of placing a verifier in a
context where it has already carried out the kind of repair it is about to be asked to judge}. Its
false-alarm rate falls by up to $11.5\pp$ with no gain in discrimination we can detect, and a metric
computed on flagged items alone will read that as an improvement.

Two recent results point opposite ways, and neither isolates the manipulation. \citet{jin2026} find
that prompt formats demanding explanations and fixes \emph{increase} misjudgement of correct code, but
there the fix is requested in the same response as the audit, so the effect is confounded with output
format. \citet{khullar2026} find monitors going easier on work framed as their own, yet report that
``explicitly stating that the action comes from the monitor does not by itself induce self-attribution
bias'': their effect rides on \emph{where} the work sits, not whose it is said to be.

Our design separates what those conflate. The present task (problem text, step text, instructions,
output schema, evidence budget) is \textbf{byte-identical} across conditions, enforced by tests that
diff rendered prompts, so nothing here is an output-format effect; and the audit\,$\to$\,repair context
is a completed exchange about a \emph{different} item, never a rewording of the request being answered.

The dependent variable is the \textbf{false-alarm rate} (FAR): how often a model reports an error in a
trace human annotators verified as correct. FAR rather than accuracy, because the manipulation is
hypothesised to move a \emph{threshold}, and because each false alarm costs a repair cycle on something
already right.

\paragraph{Contributions.}
\begin{enumerate}[topsep=2pt,itemsep=1pt,leftmargin=*]
\item \textbf{The effect} (\S\ref{sec:effect}). A prior audit\,$\to$\,repair episode lowers false
  alarms in 15 of 15 model\,$\times$\,wording combinations against a length-matched filler, surviving
  five controls.
\item \textbf{Polarity drift alone cannot explain it} (\S\ref{sec:wedge}). An episode reporting an \emph{error}
  lowers false alarms \emph{further}, at all five wordings on the model where that cell is clean,
  which is the opposite sign to what the accumulated-message account predicts. Decomposing further,
  repair content and the audit's verdict are complementary across models, so no single component is
  necessary.
\item \textbf{What it does to the instrument} (\S\ref{sec:criterion}). The change is a criterion shift
  with no discrimination gain we can detect, and a hand audit of the false alarms it removes shows the
  shift is nonetheless beneficial at this operating point.
\end{enumerate}

\section{Related work}
\label{sec:related}

\paragraph{Prior conversation shifts judgements the other way.}
The closest result is \citet{temkit2026}, who across 84{,}088 calls to 12 models shows that judgements
drift toward the polarity of the preceding conversation ($d = -0.17$), concentrated on items where the
model is uncertain at baseline, with a \textbf{negativity asymmetry}: negative histories induce
$1.52\times$ more drift than positive ones. That account makes a sharp prediction here. Our
incorrect-verdict cell places an audit reporting an \emph{error} in the context (an unambiguously
negative history) and so should raise the false-alarm rate. It lowers it, and by more than a clean
audit does (\S\ref{sec:wedge}). That work also reports the drift neither growing with context length (5 prior turns and 50 give
the same shift) nor varying with position, so neither is available to explain a \emph{larger} effect.

\paragraph{Verifier strictness as a manipulable quantity.}
\citet{verifier2026} steer verifier strictness directly on ProcessBench, the same benchmark,
establishing it as a movable axis. What they do not ask is whether an ordinary pipeline arrangement
moves it with nobody intending to.

\paragraph{Judges, and what makes them move.}
LLM-as-judge evaluation is now standard \citep{zheng2023judging,gu2024survey}, with documented
sensitivity to self-preference \citep{panickssery2024llm}, to sycophantic agreement with a user
\citep{sharma2023sycophancy}, to the ordering of in-context examples \citep{lu2022fantastically}, to provided knowledge
\citep{li2025} and to paraphrase \citep{judgesense2026}. Ours is a different lever: not who the judge
is talking to or how the question is worded, but what job the judge has been told comes next.

The critique\,$\to$\,correction pipeline itself is benchmarked by CriticBench \citep{lin2024} and
CriticEval \citep{lan2024}, and \citet{yang2025} decompose self-correction into confidence and critique
components; self-correction work more broadly asks whether models can repair their own output
\citep{huang2024large,olausson2024self}. All of these measure how well the critique and the correction
are done. We ask something upstream of that: what a completed repair already in the context does to the
critique itself.

\paragraph{Signal detection.}
A false-alarm rate on its own cannot separate a model that discriminates better from one that has
become reluctant to flag. We report $d'$ and the criterion $c$ \citep{macmillan2004detection},
following recent applications of signal-detection theory to language models \citep{sdt2026}, and treat
the pairing of a false-alarm rate with a detection rate on labelled-incorrect traces as the minimum
needed to interpret any FAR movement at all.

\section{Method}
\label{sec:method}

\begin{figure*}[t]
\centering
\includegraphics[width=0.85\textwidth]{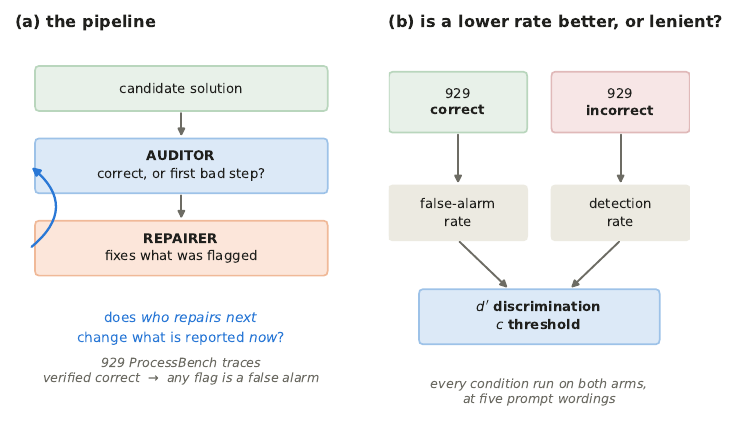}
\caption{The question and the instrument. \textbf{(a)} An auditing pipeline: a checker reports, a
repairer fixes. We ask whether \emph{who repairs next} changes what is reported \emph{now}, on traces
human annotators verified correct, so any flag is a false alarm. \textbf{(b)} A false-alarm rate alone
cannot distinguish better discrimination from greater reluctance to flag, so every condition is run on
both a correct and a labelled-incorrect arm and combined into $d'$ and the criterion $c$. The two arms
are disjoint and matched trace for trace on source; the prior-context conditions use a
source-proportional half of the correct arm (465 targets), the ladder all 929.}
\label{fig:design}
\end{figure*}

\subsection{Task and dependent variable}

Each item is a mathematical problem with a step-by-step candidate solution. The model returns one JSON
object under a constrained decoding schema: verdict, first-error step, confidence, error type, and at
most forty words of evidence (Appendix~\ref{app:prompts}). Reasoning traces are disabled deliberately,
because variable-length thinking would break the identical-budget invariant the comparison rests on;
\S\ref{sec:robustness} turns them back on.

Items come from ProcessBench \citep{zheng2024processbench}, using only traces whose label marks every
step correct, so \emph{a reported error is a false alarm by construction}. From an eligible pool of
1{,}101 we allocate three disjoint, source-stratified arms: \textbf{929 clean} targets, \textbf{50
warmup} items for generating episodes, and 122 held in reserve. Disjointness is load-bearing: an
episode item that was also a target would mean the model had already audited the very trace it is
later asked to judge cold.

The signal side of the signal-detection analysis (\S\ref{sec:criterion}) is a second, disjoint arm
of \textbf{929 labelled-incorrect} traces drawn with the same seed and matched to the clean arm's
source mix trace for trace (126 GSM8K, 315 MATH, 301 OlympiadBench, 187 OmniMath); the two arms share
no item. The incorrect-verdict episodes (AX, AXN) come from a further 50 labelled-incorrect traces per
model, source-stratified and verified to overlap neither the 929 detection targets nor the
1{,}101-trace correct pool. ``Detection rate'' throughout means the rate at which a model returns the
\texttt{incorrect} verdict on those traces, not whether it localises the faulty step.

FAR is measured from \textbf{8 samples per item at $T{=}0.7$}. Sampling is required, not incidental: at
$T{=}0$ each item contributes a hard $0/1$, so a manipulation that shifts every item's report propensity
a few points while flipping few argmaxes would be invisible.

We audit three models (Qwen3.6-27B, Qwen3.6-35B-A3B and Ministral-3-14B), chosen by a
pre-specified instrument-sensitivity screen that a further two candidates failed
(Appendix~\ref{app:screen}): it asks whether a model's FAR responds to explicit framing \emph{at all},
in either direction, since a dependent variable that does not move cannot inform a null.
Every condition is measured under \textbf{five semantically matched
wordings} (F1--F5) of the audit instruction (\S\ref{sec:robustness}).

\subsection{The prior-context conditions}
\label{sec:design}

The manipulation places a completed audit\,$\to$\,repair episode on a held-out item into the context
before the target request, on a source-proportional half of the clean arm (465 of 929 targets), with
the no-context baseline \textbf{R0} re-measured inside that subset. Episodes are generated by each
auditor itself at $T{=}0$ and then frozen, so the episode is genuinely that model's own work and is
byte-identical across cells.

Attributing an episode to another model requires either relabelling an assistant turn (a
\emph{stated} attribution, which \citet{khullar2026} report is null) or moving it into a user turn,
which reintroduces their placement confound. Neither alone is interpretable, so we cross them into a
$2\times2$: \textbf{AS} (assistant turn, self), \textbf{AO} (assistant, labelled peer), \textbf{US}
(user, labelled self), \textbf{UO} (user, peer). Labels are length-matched, and tests enforce that
stripping them makes all four cells textually identical. Throughout, ``episode'' in a contrast means the
\textbf{mean of those four cells}; a contrast naming AS means AS.

\subsection{Five controls}
\label{sec:controls}

The $2\times2$ was close to flat on the two Qwen models (all four cells within $0.63$ and
$0.72\pp$ of each other) but not on Ministral, whose cells span $5.65\pp$ because that is the one
model with a placement effect (Appendix~\ref{app:ladder}). What all four share and R0 lacks is
\emph{a prior exchange at all}, which the $2\times2$ cannot distinguish from the episode's content.
Each control below removes one more alternative.

\begin{description}[topsep=2pt,itemsep=1pt,leftmargin=*,font=\normalfont\bfseries]
\item[AF, length-matched filler.] A prior exchange on a non-audit task, in the
  same turns, paired to the same targets, length-matched \emph{per episode}
  (Appendix~\ref{app:prompts}). Separates ``the episode was an audit'' from ``there was context''.
\item[AV, audit only.] The episode with the repair request and repair deleted. AV is \emph{shorter}
  than AF, so AS $-$ AV varies whether a continuation exists as well as what it contains, which is
  why we rest nothing on it.
\item[AN, inert continuation.] The repair request replaced by ``restate, without changing it, what your
  audit concluded'', and the repair by the model's own restatement, generated in the episode's own
  context at $T{=}0$. AS and AN both carry an audit \emph{and} a second assistant turn, differing only
  in what that turn does, which is what makes AS $-$ AN identified where AS $-$ AV is not. The
  restatements came out 20--35\% \emph{longer} in mean characters than the repairs they replace, so the
  contrast is conservative with respect to length.
\item[AX, incorrect verdict.] The same structure on a ProcessBench trace \emph{labelled incorrect}, so
  the model reports an error and produces a real correction. This control exists because of an
  asymmetry in the episodes: on the Qwen models 88\% and 86\% of ordinary episodes
  report \texttt{correct} (Ministral, 34\%, is the exception), and a clean verdict leaves nothing to
  correct, so those cells place one or two \texttt{"verdict": "correct"} assertions in front of a model
  about to emit that field. AF cannot separate that from repair content: it deletes verdict and
  repair together.
\item[AXN, inert continuation on the fault.] AN's construction on AX's pool.
  This makes AX $-$ AXN the only contrast where the thing removed is a \emph{real repair of a real
  fault}. AXN $-$ AN then compares an error-verdict episode with a clean-verdict one having removed the
  repair from both sides. Note what it does \emph{not} isolate: the two sides draw on different pools,
  so the episode's trace varies with its verdict, and separating the two would need a fourth pool we
  have not run.
\end{description}

\noindent The resulting decomposition, with what each contrast holds fixed:

\begin{center}\small
\begin{tabular}{@{}l>{\raggedright\arraybackslash}p{0.46\columnwidth}@{}}
\toprule
contrast & what varies \\
\midrule
episode $-$ AF & audit content, length held \\
AV $-$ AF & audit alone vs non-audit \\
AN $-$ AF & audit $+$ inert turn vs non-audit \\
\textbf{AS $-$ AN} & continuation content only \\
\textbf{AX $-$ AXN} & content only, on a real fault \\
\textbf{AXN $-$ AN} & error vs clean episode, repair removed from both \\
\textbf{AX $-$ AS} & polarity \emph{and} repair reality \\
AS $-$ AV & presence \emph{and} content (not identified) \\
\bottomrule
\end{tabular}
\end{center}

Under polarity drift \citep{temkit2026}, an episode whose audit reports an error should push FAR
\emph{up} relative to one reporting none. Under experienced repair it should behave like AS. The two
make opposite directional predictions, which is what makes AX decisive.

\subsection{Statistical conventions}
\label{sec:stats}

Three choices, each load-bearing.

\textbf{Intervals cluster on the reused episode.} All intervals are 20{,}000-replicate bootstraps on
per-item paired differences. Each prior-context cell cycles 465 targets over a pool of 50 frozen
episodes, so \textbf{each episode is reused for about 9.3 targets} and two targets sharing one see the
same prior exchange. Every such interval is therefore a cluster bootstrap over episodes, which puts the
effective sample size for the episode-attributable component nearer 50 than 465, widening the headline
contrast's interval by a median of $14\%$ against item resampling (over all claim contrasts $7\%$, and
narrower in about a quarter; both are persisted). Calibration was verified by simulation at this
study's geometry (coverage 0.94 against a nominal 0.95). $d'$, $c$ and balanced accuracy pair a
false-alarm rate on 465 targets with a detection rate on 929, so no per-item pairing exists across the
two: each arm is resampled independently and clustered on its own episode assignment (9.3 targets per
episode on the clean side, 18.6 on the incorrect side), and the two draws are combined per replicate.

\textbf{One significance criterion.} Every resampled $p$ is the achieved significance level of the
interval printed beside it, so the two cannot disagree. The hand audit alone is not resampled; its
exact binomial intervals and Fisher test are named where they appear.

\textbf{Multiplicity over families declared in code.} Holm--Bonferroni step-down within families fixed
before any $p$-value was visible, since a family chosen afterwards is not a correction. The claim
family is the eight contrasts of \S\ref{sec:controls} across three models, $k=24$; the $2\times2$
factorial contrasts form their own family, $k=9$; and $\Delta d'$, $\Delta c$ and $\Delta$balanced
accuracy for the three prior-context cells (AF, AS, AV) across three models form one family each,
$k=9$. The last two matter: the paper
concludes from criterion counts as well as from $d'$ counts. Each wording is corrected as its own
set of families over the same declared membership, so cross-wording summaries are counts, never
$p$-values. \holm{} marks survival of the declared family and is the only mark we draw a conclusion
from.

\section{A prior audit--repair episode lowers false alarms}
\label{sec:effect}

Against the length-matched filler the effect is $-4.00$, $-3.59$ and $-8.83\pp$ at our primary wording
(Table~\ref{tab:main}), and it holds in \textbf{all 15} model\,$\times$\,wording combinations, ranging
$-2.8$ to $-11.5\pp$, every $p$ at the $20{,}000$-replicate floor after episode clustering. This is the result that
survives every control we applied.

The filler is what makes it a claim about audits rather than about context. Taking the models in
Table~\ref{tab:main}'s column order, AF $-$ R0 is $+1.58$, $-0.18$ and $-0.30\pp$
($p = 0.0045$, $0.86$, $0.71$): null on two of three, and on Qwen3.6-27B it moves \emph{opposite} to the
episode, which makes that model's episode $-$ R0 figure conservative rather than inflated. Prior context
of any kind does not do this; prior context that was an audit does.

Two further readings from Table~\ref{tab:main}, and both need the full sweep
(Table~\ref{tab:wordings}) rather than F1 alone. At F1 the repair \emph{request} survives on
\textbf{all three} models --- AS $-$ AN is $-1.31$, $-2.35$ and $-3.13\pp$, every one clearing $k=24$
--- while the audit with an inert turn in place of the repair (AN $-$ AF) survives on \textbf{one}.
Across all five wordings that ordering reverses: AN $-$ AF survives \textbf{12 of 15} and AS $-$ AN
\textbf{10 of 15}. So neither component is dispensable and neither is safe to read at one wording,
which is the general point of \S\ref{sec:robustness}. Merely having
a continuation contributes on none (AV $-$ AN: $-0.10$, $-0.69$, $-0.63\pp$, all $p > 0.2$).

It also cuts against a simple instruction-following account. On Qwen3.6-35B-A3B the episode moves FAR
by $-3.59\pp$ while a \emph{maximal explicit leniency prime} on that model moves it only
$-1.76\pp$ \ci{-3.72}{+0.07}, an interval touching zero, so not an established effect at all
(Appendix~\ref{app:screen}): whatever a prior audit does to this model, an explicit instruction
does not reach.

\begin{table*}[t]
\centering\small
\begin{tabular}{@{}lrrr@{}}
\toprule
& Qwen3.6-27B & Qwen3.6-35B-A3B & Ministral-3-14B \\
\midrule
R0 false-alarm rate & 0.185 & 0.232 & 0.691 \\
\midrule
\multicolumn{4}{@{}l}{\emph{Is it an audit, or merely context?}} \\
episode $-$ AF & $-4.00$\holm{} \ci{-5.26}{-2.80} & $-3.59$\holm{} \ci{-5.38}{-1.92} & $-8.83$\holm{} \ci{-11.83}{-5.91} \\
AF $-$ R0 (the filler itself) & $+1.58$ \ci{+0.46}{+2.71} & $-0.18$ \ci{-2.12}{+1.69} & $-0.30$ \ci{-1.83}{+1.18} \\
\midrule
\multicolumn{4}{@{}l}{\emph{Which part of the episode?}} \\
AV $-$ AF \, (audit alone) & $-3.13$\holm{} \ci{-4.38}{-1.98} & $-1.62$ \ci{-3.48}{+0.19} & $-3.67$ \ci{-6.39}{-0.96} \\
AN $-$ AF \, (audit $+$ inert turn) & $-3.03$\holm{} \ci{-4.27}{-1.85} & $-0.93$ \ci{-2.72}{+0.79} & $-3.04$ \ci{-5.97}{-0.14} \\
AS $-$ AN \, (repair request) & $-1.31$\holm{} \ci{-1.97}{-0.73} & $-2.35$\holm{} \ci{-3.81}{-0.88} & $-3.13$\holm{} \ci{-5.41}{-0.90} \\
AV $-$ AN \, (continuation presence) & $-0.10$ \ci{-0.91}{+0.89} & $-0.69$ \ci{-1.85}{+0.42} & $-0.63$ \ci{-2.56}{+1.43} \\
\bottomrule
\end{tabular}
\caption{The episode effect and its controls at wording F1, 465 targets, FAR percentage points;
negative means \emph{fewer} false alarms. Every interval is a 95\% cluster bootstrap on the frozen
episode (\S\ref{sec:stats}). ``episode'' is the mean of the four $2\times2$ cells, not AS alone.
\holm{} marks survival of Holm--Bonferroni within the declared claim family ($k=24$), the only mark we
draw a conclusion from. Ministral's AV $-$ AF and AN $-$ AF are therefore shown with intervals
excluding zero that we nonetheless do not claim.}
\label{tab:main}
\end{table*}

\section{Polarity drift alone cannot explain it}
\label{sec:wedge}

\subsection{The decisive contrast has the wrong sign}

If a prior audit lowers false alarms because the model drifts toward the polarity of its context, then
an audit reporting an \emph{error} (the negative pole, and the one \citet{temkit2026} finds
$1.52\times$ stronger) must raise them. It does the opposite. On Ministral-3-14B, AX $-$ AS is
$-5.62\pp$ \ci{-8.27}{-2.87} at our primary wording and negative at \textbf{all five}, ranging $-4.05$
to $-5.90\pp$ with every one surviving its declared family (Table~\ref{tab:decomp}). An episode in
which the model found and fixed a real error lowers subsequent false alarms \emph{more} than one in
which it found nothing: the sign opposite to the prediction, at all five wordings. Against R0
the faulty-episode cell moves FAR by $-3.27$, $-7.30$ and $-12.09\pp$ across the three models. A
within-pool split of Ministral's own episodes by the verdict they reached converges with this and needs
no extra cell (Appendix~\ref{app:withinpool}).

The other two models do not carry the test, and the reason is measurable rather than mysterious. The
manipulation only lands if the faulty episode actually contains a correction, and on Qwen3.6-27B
\textbf{37 of 50} of those ``repairs'' still end on a \texttt{"verdict": "correct"} assertion; its
contrast is null at every wording. On Qwen3.6-35B-A3B the effect is large at F1 ($-3.83\pp$
\ci{-5.51}{-2.23}) but does not reach its family at the other four, so we count the wedge as one model
of three rather than reading a single wording.

What AX $-$ AS rules out is polarity drift and verdict echoing. It does not rule out broader semantic
priming or base-rate calibration, because AX varies content and difficulty as well as verdict. The
content-matched verdict flip that would settle it needs a different design rather than more compute:
forcing the verdict would break the invariant that episodes are the model's own work.

\subsection{No single component is necessary}

Three components separate: what the repair \emph{content} adds (AX $-$ AXN, both sides carrying a
genuine \texttt{incorrect} audit), what an \emph{error-verdict episode} does once the repair is removed
from both sides (AXN $-$ AN), and what the repair \emph{request} elicits (AS $-$ AN). Measured at all
five wordings (Table~\ref{tab:decomp}), they are \textbf{complementary rather than consistent}:

\begin{itemize}[topsep=2pt,itemsep=1pt,leftmargin=*]
\item Repair content carries the two Qwen models, surviving at \textbf{4 of 5} wordings each, and is
  null on Ministral at all five.
\item The error-verdict episode is the mirror image: \textbf{5 of 5} on Ministral, where it is the
  largest single component at $-8.96\pp$, against 1 of 5 on the 35B and 0 of 5 on Qwen3.6-27B.
\item The repair \emph{request} is the only component with survivors on all three models.
\end{itemize}

Each component therefore fails to survive on precisely the model the other explains, which is why we
name no mechanism. What the evidence supports is that different components account for the effect
across models, both in the same direction, and that we do not identify a single common route.

\begin{table*}[t]
\centering\footnotesize
\begin{tabular}{@{}llrrrrrc@{}}
\toprule
model & contrast & F1 & F2 & F3 & F4 & F5 & surv. \\
\midrule
Qwen3.6-27B & AS $-$ AN \, (repair request) & $-1.31$$^{\dagger}$ & $-0.85$$^{\dagger}$ & $-0.92$$^{\dagger}$ & $-0.39$ & $-0.54$ & 3/5 \\
Qwen3.6-35B-A3B &  & $-2.35$$^{\dagger}$ & $-1.90$$^{\dagger}$ & $-2.42$$^{\dagger}$ & $-1.80$$^{\dagger}$ & $-1.92$$^{\dagger}$ & 5/5 \\
Ministral-3-14B &  & $-3.13$$^{\dagger}$ & $-2.15$ & $-3.10$ & $-4.03$$^{\dagger}$ & $-2.11$ & 2/5 \\
\midrule
Qwen3.6-27B & \textbf{AX $-$ AS} \, (wedge) & $-0.51$ & $+0.02$ & $+0.99$ & $-0.81$ & $-0.50$ & 0/5 \\
Qwen3.6-35B-A3B &  & $-3.83$$^{\dagger}$ & $-0.80$ & $-1.27$ & $-0.17$ & $-1.18$ & 1/5 \\
Ministral-3-14B &  & $-5.62$$^{\dagger}$ & $-4.05$$^{\dagger}$ & $-5.90$$^{\dagger}$ & $-5.16$$^{\dagger}$ & $-4.33$$^{\dagger}$ & 5/5 \\
\midrule
Qwen3.6-27B & AX $-$ AXN \, (repair content) & $-1.46$$^{\dagger}$ & $-1.32$$^{\dagger}$ & $-0.43$ & $-1.74$$^{\dagger}$ & $-1.17$$^{\dagger}$ & 4/5 \\
Qwen3.6-35B-A3B &  & $-2.84$$^{\dagger}$ & $-1.35$ & $-2.69$$^{\dagger}$ & $-1.88$$^{\dagger}$ & $-1.49$$^{\dagger}$ & 4/5 \\
Ministral-3-14B &  & $+0.22$ & $-0.63$ & $+0.00$ & $-1.33$ & $-0.31$ & 0/5 \\
\midrule
Qwen3.6-27B & AXN $-$ AN \, (error episode) & $-0.36$ & $+0.50$ & $+0.51$ & $+0.54$ & $+0.13$ & 0/5 \\
Qwen3.6-35B-A3B &  & $-3.34$$^{\dagger}$ & $-1.35$ & $-0.99$ & $-0.09$ & $-1.61$ & 1/5 \\
Ministral-3-14B &  & $-8.96$$^{\dagger}$ & $-5.56$$^{\dagger}$ & $-9.00$$^{\dagger}$ & $-7.86$$^{\dagger}$ & $-6.12$$^{\dagger}$ & 5/5 \\
\bottomrule
\end{tabular}
\caption{The 4 claim contrasts discussed in the text at every wording, all three models, FAR percentage points; negative means fewer false alarms. Each interval is a 95\% cluster bootstrap on the frozen episode, and $^{\dagger}$ marks survival of Holm--Bonferroni within that wording's claim family ($k=24$), corrected per wording over the same declared membership. The `surv.' column is the count the body quotes. Intervals are omitted for width; they are in the released artefacts.}
\label{tab:decomp}
\end{table*}

\section{A threshold move, and a beneficial one}
\label{sec:criterion}

A lower false-alarm rate is not by itself good news. A model that has become reluctant to flag anything
will show one, and so will a model that has genuinely got better at telling correct traces from faulty
ones. Separating those requires the second arm: the same conditions on 929 ProcessBench traces
\emph{labelled incorrect}, giving a detection rate to pair with each false-alarm rate.

\begin{figure*}[t]
\centering
\includegraphics[width=\textwidth]{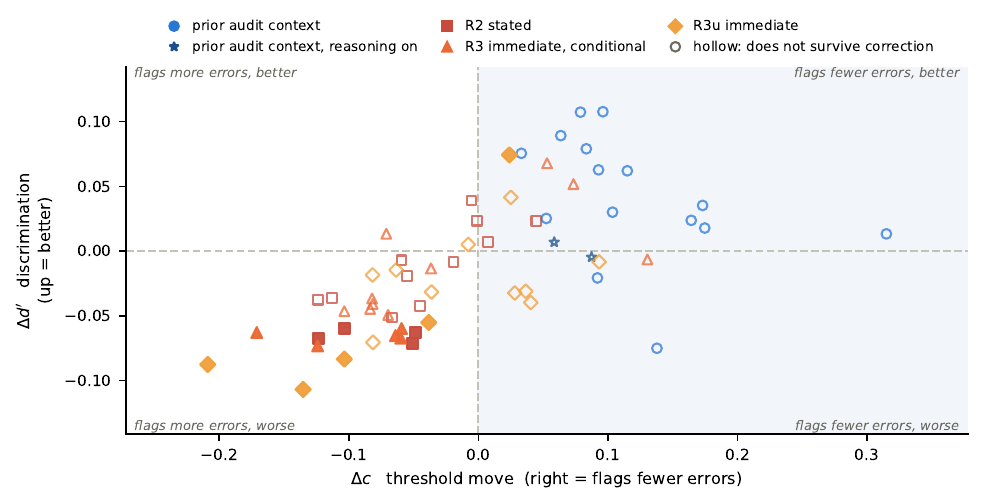}
\caption{Every detection-arm contrast as one point in (threshold move, discrimination change), all
three models and all five wordings. \textbf{Fill encodes survival of the $\Delta d'$ family only}, so
no prior-context point is filled: that is the $0$ of $15$ of \S\ref{sec:criterion}, read off the
figure. Quadrants are labelled by what the auditor visibly does, because signal-detection usage and
this paper's usage point opposite ways for the same direction: a higher criterion is ``conservative''
about asserting an error, which is leniency toward the audited work. The prior-context cell sits right
of zero and, in all but two cases, just above the axis; the stars are the same cell with reasoning on
(\S\ref{sec:robustness}). Of the 14 filled points, 13 are the Qwen prospective rungs of
Appendix~\ref{app:ladder}, all below zero; the fourteenth is Ministral's R3u, the one survivor in the
upper-right quadrant.}
\label{fig:quadrant}
\end{figure*}

\paragraph{The threshold moves; no detectable gain in discrimination.}
The criterion $c$ moves the same way, toward flagging less, in \textbf{15 of 15}
model\,$\times$\,wording combinations, and survives its declared family in \textbf{13}. Sensitivity
does not keep up: $\Delta d'$ survives in \textbf{0 of 15}. Two things stop that being a claim that
discrimination is unchanged. The $\Delta d'$ estimates are positive in all but two of the 15, leaning
the same way as the criterion; and because the two arms are resampled independently,
$\mathrm{SE}(\Delta d')$ is exactly twice $\mathrm{SE}(\Delta c)$, so a sensitivity effect must be
double the size of a criterion one to clear the same threshold. The median ratio of $|\Delta c|$ to
$|\Delta d'|$ in the same combination is $1.85$. Both are measured against R0 rather than the filler,
which on the 27B moves opposite to the episode and so leaves that estimate conservative. The threshold
moved, and any sharpening is smaller
than this design resolves --- which is not the same as none.
Figure~\ref{fig:quadrant} shows the separation directly.

\paragraph{The operating point still improves.}
Balanced accuracy, the average of the detection rate and one minus the false-alarm rate, rises in
\textbf{15 of 15} combinations and survives the same correction in 6, including $+1.22\pp$
\ci{+0.38}{+2.14} on the 27B and $+2.62\pp$ \ci{+1.14}{+4.10} on Ministral at F1. The rates behind
those, AS against R0 at F1: detection on the incorrect arm $0.908\!\rightarrow\!0.905$,
$0.901\!\rightarrow\!0.885$ and
$0.969\!\rightarrow\!0.957$, against clean-arm false-alarm reductions of $2.8$, $3.5$ and $6.5\pp$. Detection
\emph{does} fall ($0.3$, $1.6$ and $1.2\pp$), but the false-alarm side moves several times further, so
balanced accuracy still rises.

\paragraph{Are these false alarms real errors?}
This assumption carries the whole interpretation, so we measured it on the baseline cell.
Two of the cells were re-run capturing the full audit JSON, and we hand-audited a sample of
50 \textbf{R0} false alarms from Ministral at F1 against the rubric of Appendix~\ref{app:rubric}. \textbf{41 (82\%) are simply
wrong}, 8 (16\%) are defensible-but-stricter readings, 1 (2\%) is a suspected gold-label error, and none
were unclear. The 16\% carries a Wilson interval of \ci{\text{8.3\%}}{\text{28.5\%}}. One specimen
conveys the majority better than any aggregate, flagging a step
while conceding it: \emph{``5 trees $\times$ 6 lemons/year $\times$ 10 years = 300 is correct, but the
unit isolation is logically inconsistent''}.

Defensible flags are not spread evenly. They concentrate in the \texttt{logical} error type (7/20) and
we found \textbf{none in 21 arithmetic or algebraic cases}, a Clopper--Pearson interval of
\ci{\text{0\%}}{\text{16.1\%}} pooled, so not an absence. Only the pooled comparison survives (Fisher exact $p = 0.0034$); neither type alone
survives correction over ten pairwise comparisons. Decomposing the FAR reduction by error type
accordingly, arithmetic and algebraic flags account for 97\% of it ($-3.85$ and $-2.40$ of the
$-6.47\pp$ AS $-$ R0 reduction on that model and wording), which is where the fabricated flags are.

Taken together: the episode makes these models more lenient without a detectable gain in discrimination,
and because four in five of its baseline false alarms are fabrications, leniency is the right
direction \emph{at this operating point}.

\section{Robustness}
\label{sec:robustness}

\paragraph{Wording.}
Table~\ref{tab:decomp} is the sweep for the four claim contrasts and
Table~\ref{tab:wordings} has all eight; running it was not a formality: our
pre-specified wording was unrepresentative more than once, and no contrast here is safe to read at a
single wording.

\paragraph{Reasoning traces.}
The main study disables reasoning to hold the output budget identical, which removes the mechanism
deployed judges actually use. We re-ran R0, AS and AF with thinking enabled on both Qwen
models (22{,}320 generations; the Mistral models do not expose the flag). Reasoning makes these
models much better auditors (R0 FAR $0.185\!\rightarrow\!0.063$ and $0.232\!\rightarrow\!0.059$), and
the filler-controlled effect survives on \textbf{both}: AS $-$ AF is $-1.30\pp$ \ci{-1.96}{-0.70} and
$-1.10\pp$ \ci{-1.81}{-0.43}, which against each arm's own baseline is $-19.7\%$ and $-17.5\%$ with
thinking on versus $-21.6\%$ and $-14.3\%$ with it off: the proportional effect is preserved and only
the absolute margin shrinks with the baseline. The arm costs $\approx18\times$ the output tokens, and
truncation is not zero (7--27 of 3{,}720 choices per cell) with dropout outcome-associated rather than
random. Thinking and \texttt{max\_tokens} both differ from the main study, so only within-arm
contrasts are interpretable.

\paragraph{And the threshold reading holds there.}
Running the second arm with thinking on is what lets $c$ and $d'$ be computed in that setting. Against
R0 at F1, $\Delta c$ moves away from flagging on both models ($+0.087$ \ci{+0.052}{+0.125} and
$+0.058$ \ci{+0.016}{+0.102}) while $\Delta d'$ sits on zero ($-0.0046$ \ci{-0.076}{+0.068} and
$+0.0069$ \ci{-0.079}{+0.092}), without even the positive lean of \S\ref{sec:criterion}. Balanced
accuracy does \emph{not} rise here (all four $p \ge 0.16$), so that section's benefit belongs to its
operating point rather than to this one. One wording, two models: a check on \S\ref{sec:criterion},
not a second claim family.

\paragraph{Reproducibility.}
The dependent-variable audit re-ran two cells three months later on the same frozen pools: AS
reproduced its false-alarm rate \emph{exactly} ($0.625960$) and R0 to five decimals. Pools are
content-hashed and contrasts refuse to cross two hashes, because $T{=}0$ is not reproducible under
continuous batching.

\paragraph{Prospective responsibility, in brief.}
A ladder of merely \emph{stated} future obligations behaves differently from the experienced episode and
splits by model family, so we report it in Appendix~\ref{app:ladder} rather than claiming it here. One
thing from it carries over: every prospective rung that moves $d'$ on the Qwen models moves it
\emph{down} (13 of 13 surviving contrasts), so a stated repair role is not a free improvement.

\section{Conclusion}

How a checking pipeline is wired changes what the checker reports. A completed
audit\,$\to$\,repair episode in the context lowers false alarms in every model\,$\times$\,wording
combination we measured, by up to $11.5\pp$, and where the manipulation lands cleanly it does so in the
direction opposite to what prior-context polarity drift predicts: an episode that reported an error is
more lenient still, not less. The change is a threshold move with no discrimination gain we can detect,
and because four in five of its baseline false alarms are fabrications, it happens to help at this
operating point. That last clause is the part a practitioner should not rely on: the threshold moved
without anyone asking it to, and whether a threshold move helps depends on an operating point a
pipeline change can silently alter.

\section*{Limitations}
\label{sec:limitations}

The wedge holds at all five wordings on
one model of three: Qwen3.6-27B's AX cell is degenerate rather than contradictory (37/50 repairs
still assert \texttt{correct}) and the 35B reaches its family at one wording only. Because AXN and AN draw
on different pools, neither contrast separates the
verdict token from the trace it describes. The reasoning arm covers two of three models, and dropout
there --- truncation and schema-invalid output alike --- is outcome-associated rather than random. The
false-alarm audit is one model at one wording, by one author with no
inter-annotator agreement (Appendix~\ref{app:rubric}). Two further candidates were
screened out pre-hypothesis, and on one the episode contrast is significant and
\emph{reversed} ($+1.58\pp$ \ci{+0.45}{+2.82}), so our direction describes instrument-responsive
models rather than every verifier we tried (Appendix~\ref{app:screen}). All models are open-weight:
one benchmark, one task family, no frontier or closed judge.

\bibliography{refs}

\appendix

\section{The instrument-sensitivity screen}
\label{app:screen}

Before any hypothesis was tested, each candidate model was measured under three framing controls --- a
plain restatement (PC), a maximal explicit leniency prime (PCL) and a maximal strictness prime (PCH)
--- each judged against \emph{its own} sham-derived noise band rather than a pooled one. The rule, as
implemented in \texttt{analyse\_screen} and fixed before any hypothesis was tested, is:

\begin{quote}
let $b_m = \max(|\text{lo}|, |\text{hi}|)$ of model $m$'s sham interval; $m$ passes if for \emph{any}
control $c \in \{\text{PC}, \text{PCL}, \text{PCH}\}$ the interval of $c$ clears $b_m$ in either
direction, i.e.\ $\text{lo}_c > b_m$ or $\text{hi}_c < -b_m$.
\end{quote}

The criterion is responsiveness to framing \emph{at all}, not to leniency specifically, and that is
what separates the retained models from the excluded ones: Qwen3.6-35B-A3B clears its band on PC
($+6.78\pp$) and PCH ($+12.28\pp$) while PCL does not resolve ($-1.76\pp$ \ci{-3.72}{+0.07}), so it
passes upward-only; the Gemma candidates clear it on \emph{none} of the three, with a framing range of
$1.94\pp$ against the 35B's $8.54\pp$ --- a dependent variable that does not move at all.

The asymmetry still matters for reading nulls. On the 35B a downward prime does not resolve, so a
downward null on that model is weak evidence, and we report measured effects rather than a
detectability bound.

The leniency prime moves FAR by $-13.93\pp$ \ci{-16.89}{-11.14} on Ministral, $-4.07\pp$
\ci{-6.16}{-2.28} on Qwen3.6-27B, and $-1.76\pp$ \ci{-3.72}{+0.07} on Qwen3.6-35B-A3B.

Two Gemma-family candidates were screened out. Their cells were still generated, and Table~\ref{tab:allfive}
shows every candidate's headline contrast so the selection can be judged rather than taken on trust. No
claim rests on the screened-out pair and they are not members of any declared family; counting them
would raise $k$ and test the claims we do make against hypotheses we never advanced.

\begin{table}[h]
\centering\scriptsize
\begin{tabular}{@{}llr@{}}
\toprule
model & status & episode $-$ AF \\
\midrule
Qwen3.6-27B & retained & $-4.00$ \ci{-5.26}{-2.80} \\
Qwen3.6-35B-A3B & retained & $-3.59$ \ci{-5.38}{-1.92} \\
Ministral-3-14B & retained & $-8.83$ \ci{-11.83}{-5.91} \\
\midrule
Gemma-4-26B & screened out & $+0.41$ \ci{-1.08}{+1.98} \\
Gemma-4-31B & screened out & $+1.58$ \ci{+0.45}{+2.82} \\
\bottomrule
\end{tabular}
\caption{All five candidates at F1, FAR percentage points. The retained rows are confirmatory and the
only ones any claim rests on; the screened-out pair is exploratory, having moved on none of the three
framing controls. The selection effect is real and visible here: on Gemma-4-31B the contrast is significant
and \emph{reversed}, which is why reading its other contrasts would be a mistake and why the direction
we report is a property of framing-responsive verifiers rather than of every model we tried.
Confirmatory rows come from the F1 artefact of record; the exploratory rows exist only in the
with-Gemma run, which differs from it in the third decimal.}
\label{tab:allfive}
\end{table}

\section{The prospective ladder}
\label{app:ladder}

Rungs of \emph{stated} rather than experienced responsibility, all with the present task byte-identical:
\textbf{R0} audit only; \textbf{R0p} a stated future task unrelated to repair, as a workload placebo;
\textbf{R1} a different model will repair using this audit; \textbf{R2} this model will repair later;
\textbf{R3} this model repairs immediately and only if the audit reports an error; \textbf{R3u} as R3
but unconditional, isolating the conditionality. Exactly one sentence varies between R0 and R3u.

On Qwen3.6-27B, conditionality is the most wording-stable effect in the study ($+2.78$ to
$+3.23\pp$ across the five), with the sign
\emph{opposite} to what work-avoidance predicts. Attribution moves FAR in 0 of 15 combinations (a
prospective replication of \citet{khullar2026}'s explicit-labelling null in a different task,
direction of time and dependent variable), and every attribution effect ($0.015$ to $0.374\pp$) falls inside its
own model's noise band, so this is a bound of roughly ``below half a percentage point'', not a
demonstration of zero. Placement survives its family on one model only, and there it is
\emph{positive}: assistant placement carries the \emph{higher} false-alarm rate, the opposite of how
self-leniency is usually read. R3u $-$ R2 is the one place our length standard is applied unevenly
(a $+48$ character, $+41\%$ mismatch); R3 $-$ R3u is matched to one character.

\section{Every claim contrast at every wording}
\label{app:wordings}

Table~\ref{tab:wordings} lists all eight, with the survival counts \S\ref{sec:effect} and
\S\ref{sec:wedge} quote.

\input{table_wordings}

\section{A within-pool verdict split}
\label{app:withinpool}
Because Ministral false-alarms often, 33 of
its 50 frozen episodes report an error on a \emph{clean} trace, so its own pool contains both
verdicts, on the same trace distribution, both genuinely the model's own output. Splitting its targets
by what their episode concluded, AS $-$ R0 is $-10.30\pp$ \ci{-13.48}{-7.04} where the episode reported
an error and $+0.77\pp$ \ci{-2.75}{+4.23} where it did not. On Qwen3.6-27B, whose wedge is null, the
same split is flat ($-2.91$ vs $-2.74\pp$). This is observational (verdicts are not assigned), so we
offer it as convergent with AX $-$ AS rather than as a test, and the Qwen splits rest on 6 and 7
error-reporting episodes respectively.

\section{The instrument}
\label{app:prompts}

The system prompt is identical in every condition:

{\footnotesize\begin{verbatim}
You are a careful mathematical reasoning
auditor. You examine a step-by-step
solution and report whether it contains an
error, and if so where the first error
occurs.

Reply with a single JSON object and nothing
else, matching this schema:
{"verdict": "correct" | "incorrect",
 "first_error_step": <1-based int or null>,
 "confidence": <number between 0 and 1>,
 "error_type": <one of ['arithmetic',
   'algebraic', 'logical',
   'misread_problem', 'unjustified_step']
   or null>,
 "evidence": "<at most 40 words>"}

Set first_error_step and error_type to null
when the verdict is correct. Keep evidence
to at most 40 words.
\end{verbatim}}

\noindent The five user-message wordings, with \texttt{\{slot\}} the position of the single varying
sentence (empty for R0) and \texttt{\{steps\}} the 1-based numbered steps:

{\footnotesize\begin{verbatim}
F1  Problem:\n{problem}\n\nSolution
    steps:\n{steps}\n\n{slot}Audit the
    solution now.
F2  Here is a problem.\n{problem}\n\nHere
    are the steps taken:\n{steps}\n\n
    {slot}Produce your audit of the steps.
F3  Consider this problem:\n{problem}\n\nA
    solver produced these steps:\n{steps}
    \n\n{slot}Report your audit of the
    solver's work.
F4  Task: audit a solution.\n\nProblem:\n
    {problem}\n\nSteps:\n{steps}\n\n
    {slot}Give your audit.
F5  Below is a problem and a proposed
    step-by-step solution.\n\n{problem}\n\n
    {steps}\n\n{slot}Now audit that
    solution.
\end{verbatim}}

\noindent The ladder rungs of Appendix~\ref{app:ladder} are exactly the sentence that fills
\texttt{\{slot\}}; R2, for instance, is ``After you submit this audit, you will be given your audit and
you will be required to produce the corrected solution.'' The screen of
Appendix~\ref{app:screen} uses the same slot: its leniency prime is ``Note that solutions of this kind
very rarely contain subtle errors, and reviewers are criticised far more for reporting an error that
turns out to be benign than for missing one.''

\paragraph{The filler, in full.} AF is the load-bearing control, so here is one verbatim. Its user turn
asks the model to restate a held-out problem at a word count computed from the audit episode it
replaces, so the length match is per-episode rather than on average (the targets run from 37 to 637
words):

\begin{quote}\small
\emph{``Restate the following problem in your own words, in about 37 words, without solving it.''}
followed by the problem text; the model's own reply is the assistant turn, e.g.\ \emph{``Ava
observes that Xavier, initially four feet tall, grew three inches, while Cole, starting at fifty
inches, grew two inches. Calculate the current height difference\ldots''}
\end{quote}

\noindent Nothing in the request or the reply names correctness, error, review or repair, which is what
makes episode $-$ AF a contrast about audit content rather than about prior context.

\paragraph{The prior-context turns, verbatim.} R0 has none of this: the target request alone. The other
three prepend a two-turn exchange about a \emph{held-out} item and then the byte-identical target
request. AS carries the audit and its repair, both the model's own work at $T{=}0$:

\begin{quote}\small
\textbf{audit} \texttt{\{"verdict": "incorrect", "first\_error\_step": 1, "confidence": 1.0,
"error\_type": "unjustified\_step", "evidence": "The solution arbitrarily assumes Z=2 without
justification\ldots"\}}

\textbf{repair request} \emph{``Now produce the corrected solution.''} \quad
\textbf{repair} the corrected steps, in the same schema.
\end{quote}

\noindent AN replaces only the second turn: the request becomes \emph{``restate, without changing it,
what your audit concluded''} and the reply is the model's own restatement. AV deletes the second turn
altogether, which is why it is shorter and why we rest nothing on AS $-$ AV. Reading the four side by
side is the identification argument: R0 varies context presence, AF varies whether the context was an
audit, AN varies whether the second turn repairs or merely restates.

\section{The false-alarm audit rubric}
\label{app:rubric}

The rubric was applied to a sample of R0 false alarms drawn \emph{proportionally} to that cell's own
\texttt{error\_type} mix (population 390 flagged items: 24.1\% arithmetic, 39.0\% logical, 18.2\%
algebraic, 14.1\% misread, 4.6\% unjustified; sample 12/20/9/7/2), so shares are reported unweighted. The four disjoint outcomes are \textbf{wrong}
(the flagged step is correct and the stated evidence does not support the flag); \textbf{defensible}
(the step is formally incomplete or under-justified, so a stricter reader could flag it);
\textbf{unclear} (neither reading is supportable from the trace); and \textbf{label\_error} (the step
is genuinely defective, so the ``false alarm'' is a true alarm on a
trace annotated error-free). The audit was performed by one author, so it carries no
inter-annotator agreement; the per-case verdicts and reasons are released so the classification can be
re-checked.

\section{Reproducibility details}
\label{app:repro}

Models, with the revision actually served: Qwen3.6-27B (\texttt{6a9e13bd}), Qwen3.6-35B-A3B
(\texttt{995ad96e}), Ministral-3-14B-Instruct-2512 (\texttt{29439f81}), and the two screened-out
candidates gemma-4-26B-A4B-it (\texttt{4d7ae498}) and gemma-4-31B-it (\texttt{842da379}). All are
open-weight and served locally with constrained JSON decoding at $T{=}0.7$, 8 samples per item,
16{,}384-token context, reasoning disabled; frozen episodes were generated at $T{=}0$. Nothing was
sent to a hosted API. Allocation seed 20260805; bootstrap seed derived per (model, contrast).

Frozen episode pools are content-hashed and every contrast verifies that both sides carry
the same hash before it is formed. Multiplicity families and the screen's membership are declared in
code rather than in prose, and the analysis prints a warning if a declared family is not fully
populated: a family missing members tests its survivors at $\alpha/k$ with a smaller $k$, which is
more lenient than declared. Per-item flags, per-item confidences and every contrast's item-only
interval alongside its clustered one are persisted, so the difference the clustering makes is auditable
rather than asserted. Code, the frozen pools, the per-case audit verdicts and all per-item outputs
accompany the submission.

\end{document}

%% file: table_wordings.tex
\begin{table*}[t]
\centering\small
\begin{tabular}{@{}llrrrrrc@{}}
\toprule
model & contrast & F1 & F2 & F3 & F4 & F5 & surv. \\
\midrule
Qwen3.6-27B & episode $-$ AF & $-4.00$$^{\dagger}$ & $-3.69$$^{\dagger}$ & $-3.66$$^{\dagger}$ & $-2.76$$^{\dagger}$ & $-3.60$$^{\dagger}$ & 5/5 \\
Qwen3.6-35B-A3B &  & $-3.59$$^{\dagger}$ & $-6.37$$^{\dagger}$ & $-6.26$$^{\dagger}$ & $-5.44$$^{\dagger}$ & $-5.15$$^{\dagger}$ & 5/5 \\
Ministral-3-14B &  & $-8.83$$^{\dagger}$ & $-7.47$$^{\dagger}$ & $-6.27$$^{\dagger}$ & $-11.50$$^{\dagger}$ & $-9.23$$^{\dagger}$ & 5/5 \\
\midrule
Qwen3.6-27B & AV $-$ AF \, (audit alone) & $-3.13$$^{\dagger}$ & $-3.15$$^{\dagger}$ & $-2.85$$^{\dagger}$ & $-2.62$$^{\dagger}$ & $-2.89$$^{\dagger}$ & 5/5 \\
Qwen3.6-35B-A3B &  & $-1.62$ & $-4.15$$^{\dagger}$ & $-5.09$$^{\dagger}$ & $-3.29$$^{\dagger}$ & $-3.38$$^{\dagger}$ & 4/5 \\
Ministral-3-14B &  & $-3.67$ & $-1.59$ & $-3.04$ & $-4.19$$^{\dagger}$ & $-2.17$ & 1/5 \\
\midrule
Qwen3.6-27B & AN $-$ AF \, (audit $+$ inert) & $-3.03$$^{\dagger}$ & $-3.12$$^{\dagger}$ & $-3.00$$^{\dagger}$ & $-2.67$$^{\dagger}$ & $-3.54$$^{\dagger}$ & 5/5 \\
Qwen3.6-35B-A3B &  & $-0.93$ & $-4.39$$^{\dagger}$ & $-4.31$$^{\dagger}$ & $-3.54$$^{\dagger}$ & $-3.25$$^{\dagger}$ & 4/5 \\
Ministral-3-14B &  & $-3.04$ & $-3.53$$^{\dagger}$ & $-2.29$ & $-5.98$$^{\dagger}$ & $-4.33$$^{\dagger}$ & 3/5 \\
\midrule
Qwen3.6-27B & AS $-$ AN \, (repair request) & $-1.31$$^{\dagger}$ & $-0.85$$^{\dagger}$ & $-0.92$$^{\dagger}$ & $-0.39$ & $-0.54$ & 3/5 \\
Qwen3.6-35B-A3B &  & $-2.35$$^{\dagger}$ & $-1.90$$^{\dagger}$ & $-2.42$$^{\dagger}$ & $-1.80$$^{\dagger}$ & $-1.92$$^{\dagger}$ & 5/5 \\
Ministral-3-14B &  & $-3.13$$^{\dagger}$ & $-2.15$ & $-3.10$ & $-4.03$$^{\dagger}$ & $-2.11$ & 2/5 \\
\midrule
Qwen3.6-27B & AS $-$ AV \, (not identified) & $-1.21$ & $-0.82$ & $-1.07$$^{\dagger}$ & $-0.44$ & $-1.19$ & 1/5 \\
Qwen3.6-35B-A3B &  & $-1.66$$^{\dagger}$ & $-2.14$$^{\dagger}$ & $-1.64$$^{\dagger}$ & $-2.05$$^{\dagger}$ & $-1.79$$^{\dagger}$ & 5/5 \\
Ministral-3-14B &  & $-2.50$ & $-4.09$$^{\dagger}$ & $-2.35$ & $-5.82$$^{\dagger}$ & $-4.27$$^{\dagger}$ & 3/5 \\
\midrule
Qwen3.6-27B & \textbf{AX $-$ AS} \, (wedge) & $-0.51$ & $+0.02$ & $+0.99$ & $-0.81$ & $-0.50$ & 0/5 \\
Qwen3.6-35B-A3B &  & $-3.83$$^{\dagger}$ & $-0.80$ & $-1.27$ & $-0.17$ & $-1.18$ & 1/5 \\
Ministral-3-14B &  & $-5.62$$^{\dagger}$ & $-4.05$$^{\dagger}$ & $-5.90$$^{\dagger}$ & $-5.16$$^{\dagger}$ & $-4.33$$^{\dagger}$ & 5/5 \\
\midrule
Qwen3.6-27B & AX $-$ AXN \, (repair content) & $-1.46$$^{\dagger}$ & $-1.32$$^{\dagger}$ & $-0.43$ & $-1.74$$^{\dagger}$ & $-1.17$$^{\dagger}$ & 4/5 \\
Qwen3.6-35B-A3B &  & $-2.84$$^{\dagger}$ & $-1.35$ & $-2.69$$^{\dagger}$ & $-1.88$$^{\dagger}$ & $-1.49$$^{\dagger}$ & 4/5 \\
Ministral-3-14B &  & $+0.22$ & $-0.63$ & $+0.00$ & $-1.33$ & $-0.31$ & 0/5 \\
\midrule
Qwen3.6-27B & AXN $-$ AN \, (error episode) & $-0.36$ & $+0.50$ & $+0.51$ & $+0.54$ & $+0.13$ & 0/5 \\
Qwen3.6-35B-A3B &  & $-3.34$$^{\dagger}$ & $-1.35$ & $-0.99$ & $-0.09$ & $-1.61$ & 1/5 \\
Ministral-3-14B &  & $-8.96$$^{\dagger}$ & $-5.56$$^{\dagger}$ & $-9.00$$^{\dagger}$ & $-7.86$$^{\dagger}$ & $-6.12$$^{\dagger}$ & 5/5 \\
\bottomrule
\end{tabular}
\caption{Every claim contrast at every wording, all three models, FAR percentage points; negative means fewer false alarms. F1--F5 are the five semantically matched wordings of the audit request, F1 being the pre-specified one. Each interval is a 95\% cluster bootstrap on the frozen episode, and $^{\dagger}$ marks survival of Holm--Bonferroni within that wording's claim family ($k=24$), corrected per wording over the same declared membership. The `surv.' column is the count the body quotes. Intervals are omitted for width; they are in the released artefacts.}
\label{tab:wordings}
\end{table*}